\documentclass{iet-ell}

\usepackage{bm}
\usepackage{color}

\setcopyright{open}
\ietVolume{00}
\ietYear{2021}
\ietDoi{ell2.10001}

\received{2 July 2021}
\accepted{To be defined}

\begin{document}

\title{Intelligent control of a single-link flexible manipulator using sliding modes and artificial neural networks}

\author[af1]{Gabriel da Silva Lima}
\orcid{0000-0001-6615-078X}
\author[af1]{Diego Rolim Porto}
\author[af1]{Adilson Jos\'e de Oliveira}
\orcid{0000-0001-6104-569X}
\author[af2]{Wallace Moreira Bessa}
\orcid{0000-0002-0935-7730}

\affil[af1]{Department of Mechanical Engineering, Federal University of Rio Grande do Norte, Natal, Brazil}
\affil[af2]{Department of Mechanical and Materials Engineering, University of Turku, Turku, Finland}

\corresp{Email: wallace.moreirabessa@utu.fi}

\begin{abstract}%
This letter presents a new intelligent control scheme for the accurate trajectory tracking of flexible link manipulators. The proposed approach is mainly based on a sliding mode controller for underactuated systems with an embedded artificial neural network to deal with modeling inaccuracies. The adopted neural network only needs a single input and one hidden layer, which drastically reduces the computational complexity of the control law and allows its implementation in low-power microcontrollers. Online learning, rather than supervised offline training, is chosen to allow the weights of the neural network to be adjusted in real time during the tracking. Therefore, the resulting controller is able to cope with the underactuating issues and to adapt itself by learning from experience, which grants the capacity to deal with plant dynamics properly. The boundedness and convergence properties of the tracking error are proved by evoking Barbalat's lemma in a Lyapunov-like stability analysis. Experimental results obtained with a small single-link flexible manipulator show the efficacy of the proposed control scheme, even in the presence of a high level of uncertainty and noisy signals.
\end{abstract}

\maketitle%

\section{Introduction}\label{sec:introduction}
Flexible link manipulators (FLMs), due to their slender profile, are usually much lighter than rigid ones and provide not only a larger payload ratio but also a higher energy efficiency \cite{Zhao2021a, Rahimi2014}. In view of these advantages, their scope of application is rapidly expanding and currently ranges from surgical tasks to space missions \cite{Ma2019,Zhao2021b}. However, the elastic behavior of flexible manipulators leads to an infinite number of underactuated degrees of freedom and, in most cases, to undesired structural vibrations \cite{Seifried2014,Qing2021}, which in fact makes the design of a control scheme for this kind of robotic device quite challenging. 

Due to their learning and approximation capabilities, artificial neural networks (ANN) have been used to deal with the inherent nonlinearities and typical uncertainties of flexible robotic manipulators \cite{Su2001, Tian2005, Liu2019}. Nevertheless, it should be highlighted that ANN alone may not guarantee the necessary robustness to allow safe operating conditions. On the other hand, by combining ANN with nonlinear control methods, the resulting intelligent controller is able to meet both stability and robustness requirements while maintaining the learning and approximation features provided by neural networks \cite{Bessa2018,Deodato2019}.

Sliding mode control (SMC), in view of its robustness property, is undoubtedly a very appealing alternative for uncertain nonlinear systems like FLMs \cite{Tang2006,Xu2018,Yang2018}. Notwithstanding the many advantages of SMC, a well-known drawback of this method is the control chattering, which in the case of FLMs can excite higher modes of vibration \cite{Alandoli2020,Meng2021}. In order to avoid undesirable chattering effects, a thin boundary layer neighboring the switching surface can be adopted, but it usually spoils the tracking performance, leading to a steady-state control error \cite{Bessa2009}. It has already been shown \cite{Tang2006,Xu2018,Yang2018,Alandoli2020,Meng2021} that the tracking error in flexible link manipulators can be reduced by combining ANN with SMC. However, it should be noted that despite ANN's ability to approximate the dynamics of flexible link manipulators and to improve the tracking performance, its architecture must be chosen very carefully, in order to avoid computational complexity and time-consuming issues related to large networks \cite{Alandoli2020}.

In this work, a new intelligent controller is proposed for the accurate trajectory tracking of single-link flexible manipulators. The control law is based on a sliding mode scheme for underactuated mechanical systems, with an adaptive neural network embedded in it to deal with the unknown dynamics of the FLM.
The boundedness and convergence properties of the closed-loop signals are rigorously proved by means of a Lyapunov-like stability analysis. 
Furthermore, the main advantages of the introduced approach can be also highlighted: 
($i$) since it is based on a robust controller for underactuated systems, the proposed scheme is able to deal with the underactuating issues straightforwardly; 
($ii$) the chosen ANN architecture requires only a single hidden layer and one neuron in the input, rather than all system states or state errors, which exponentially reduces the computational complexity of the neural network and allows its implementation in low-power microcontrollers;
($iii$) by using online learning to update the ANN weights instead of supervised offline training, the adopted neural net is able to continuously approximate the plant dynamics;
($iv$) only information related to hub angle and tip acceleration is required to be measured, which avoids the need for more sensors along the link.
The experimental results obtained with a small single-link flexible manipulator evince the effectiveness of the adopted control scheme, as well as the aforementioned features.

\section{Single-link flexible manipulator}\label{sec:model}
\begin{figure}[h]
\centering{\includegraphics[width=0.4\textwidth]{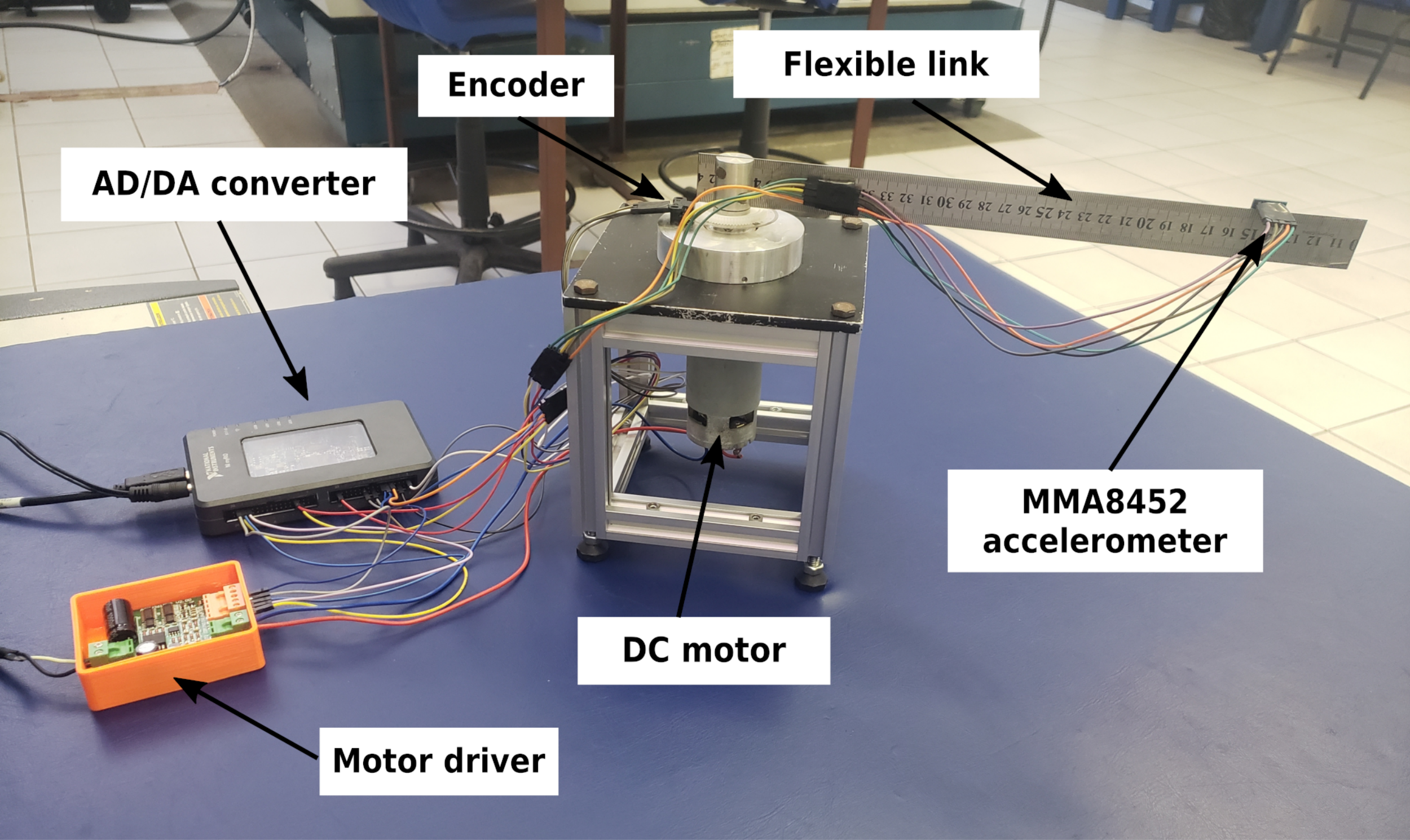}}
\caption{Experimental setup of a single-link flexible manipulator.}\label{fig:setup}
\end{figure}

Figure~\ref{fig:setup} presents the experimental setup of a small single-link flexible manipulator, designed and manufactured at the Manufacturing Laboratory of the Federal University of Rio Grande do Norte, Brazil. The device was developed to be a test bed for the evaluation of new control schemes and consists of a flexible metallic link, a brushed DC motor with gearbox, an H-bridge based motor driver, an optical incremental encoder, a low-cost MMA8452 accelerometer and an AD/DA converter (myRIO by National Instruments). 

\begin{figure}[h]
\centering{\includegraphics[width=0.29\textwidth]{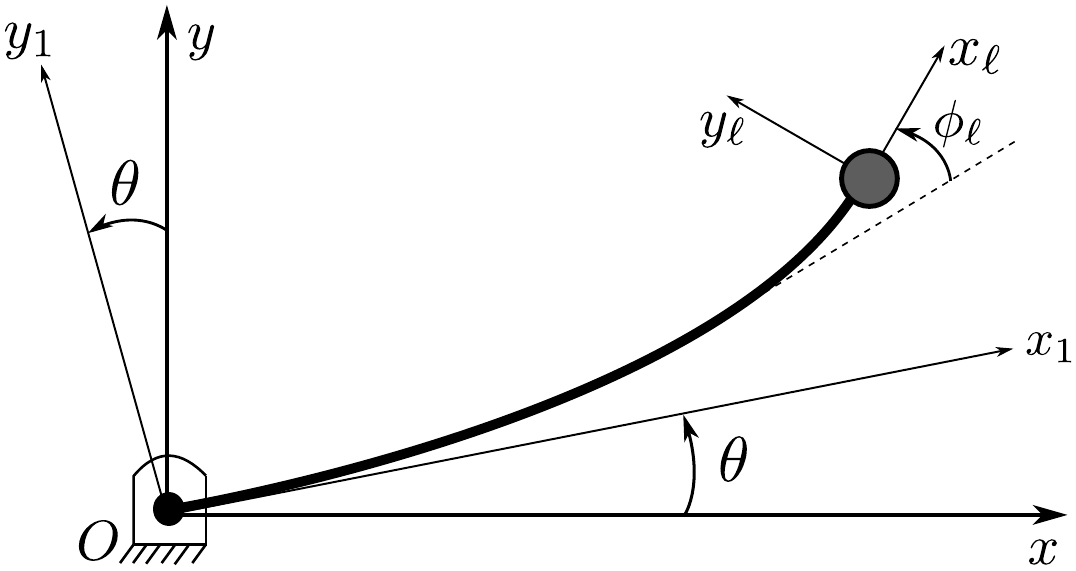}}
\caption{Schematic diagram of a single-link flexible manipulator.}\label{fig:diagram}
\end{figure}

The dynamic behavior of a flexible link manipulator can be represented by a set of coupled nonlinear partial and ordinary differential equations, which in fact represent great challenges for the design of the control law \cite{Zhao2021a,Zhao2021b}. Therefore, the adoption of any discretization technique for this kind of mechatronic system provides an effective way to solve the control problem, reducing the dynamic model to a finite dimensional system of ordinary differential equations. By using the lumped-parameter approach presented by \cite{Ge1997}, for instance, the dynamics of a single-link flexible manipulator can be expressed in the following vector form:
\begin{equation}\label{eq:link}
    \bm{M} \ddot{\bm{q}} + \bm{K}\bm{q} = \bm{b}\tau + \bm{p}
\end{equation}
where $\bm{M} \in \mathbb{R}^{m \times m}$ represents the inertia matrix, $\bm{K} \in \mathbb{R}^{m \times m}$ stands for the stiffness matrix, $\bm{b} = [1~\bm{0}]^\top$, with $\bm{0} \in \mathbb{R}^{\ell}$ being a zero vector, $\tau$ is the input torque, $\bm{p} \in \mathbb{R}^{m}$ takes all the modeling uncertainties into account, and $\bm{q} = [\theta~\phi_1~\ldots~\phi_\ell]^\top \in \mathbb{R}^{m}$ is the vector of generalized coordinates, with $m = \ell+1$ being the number of degrees of freedom, $\theta$ representing the hub angle and $\phi_i$, for $i=1,\ldots,\ell$, standing for the angular displacement of each element that composes the flexible link. Figure~\ref{fig:diagram} shows a schematic diagram of the single-link flexible manipulator with the reference frames and generalized coordinates.

Due to the lightweight and fast response of FLMs, it is usually also necessary to consider the dynamics of the actuator. Here, a first-order low-pass filter is used to represent the dynamics of the DC motor with gearbox:
\begin{equation}\label{eq:actuator}
    \dot{\tau} = - \gamma (\tau - u)
\end{equation}
where $u$ the control signal and $\gamma$ is the positive filter time constant.

\section{Intelligent controller}\label{sec:control}
In order to keep the control scheme as simple as possible so that it can be implemented on compact hardware like microcontrollers, let's consider the hub angle $\theta$ as an actuated variable and only one unactuated variable ($\ell=1$), namely the tip displacement $\phi=\phi_1$. Thus, for control purposes and taking into account the dynamics of the actuator~\eqref{eq:actuator}, the dynamic behavior of the flexible manipulator~\eqref{eq:link} can be rewritten as follows:
\begin{equation}\label{eq:flm1}
     \left[
    \begin{array}{cc}
        M_{aa} & M_{au} \\
        M_{au} & M_{uu}
    \end{array} \right]
    \left[
    \begin{array}{cc}
         \dddot{\theta}  \\
         \dddot{\phi}
    \end{array} \right] = 
     \left[
    \begin{array}{cc}
        f_{a} + \gamma u  \\
         f_{u}
    \end{array} \right] + \left[
    \begin{array}{cc}
        d_a  \\
        d_u
    \end{array} \right]
\end{equation}
where $M_{aa}$, $M_{au}$, $M_{au}$ and $M_{uu}$ are elements of the inertia matrix, $f_a$ and $f_u$ are related to the stiffness of the link, $d_a$ and $d_u$ stand for the disturbance terms, which may include external perturbations and unmodeled dynamics, and subscripts $a$ and $u$ denote, respectively, the actuated and unactuated coordinates.

By solving~\eqref{eq:flm1} for $\dddot{\theta}$ and $\dddot{\phi}$, we get:
\begin{align}
        &\label{eq:flm2a}\dddot{\theta} = M_{aa}^{\prime-1}(f_{a}^{\prime} + \gamma u + d_a^{\prime})\\
        &\label{eq:flm2b}\dddot{\phi} = M_{uu}^{\prime-1}(f_{u}^{\prime} - \gamma M_{au} M_{aa}^{-1} u + d_u^{\prime})
\end{align}
where $M_{aa}^{\prime} = M_{aa} - M_{au}^2 M_{uu}^{-1}$, $M_{uu}^{\prime} = M_{uu} - M_{au}^2 M_{aa}^{-1}$, $f_{a}^{\prime} = f_{a} - M_{au} M_{uu}^{-1} f_{u}$,  $f_{u}^{\prime} = f_{u} - M_{au} M_{aa}^{-1} f_{a}$, $d_{a}^{\prime} = d_{a} - M_{au} M_{uu}^{-1} d_{u}$, and $d_{u}^{\prime} = d_{u} - M_{au} M_{aa}^{-1} d_{a}$. 

Defining $\bm{\tilde{q}} = \bm{q} - \bm{q}_d$ as the tracking error, with $\bm{q}_d$ being the desired trajectory, and following \cite{Ashrafiuon2008}, a stable sliding surface that considers both actuated and unactuated coordinates can be defined as
\begin{equation}\label{eq:surface}
\begin{split}
    s & = \alpha_a \ddot{\tilde{\theta}} + 2\lambda_a \dot{\tilde{\theta}} + \lambda_a^2 \tilde{\theta} + \alpha_u \ddot{\tilde{\phi}} + 2\lambda_u \dot{\tilde{\phi}} + \lambda_u^2 \tilde{\phi}\\
    & = \alpha_a \ddot{\theta} + \alpha_u \ddot{\phi} + s_r
\end{split}
\end{equation}
with $s_r = - \alpha_a \ddot{\theta}_d + 2\lambda_a \dot{\tilde{\theta}} + \lambda_a^2 \tilde{\theta} - \alpha_u \ddot{\phi}_d + 2\lambda_u \dot{\tilde{\phi}} + \lambda_u^2 \tilde{\phi}$.

The intelligent controller is then proposed in order to ensure the attractiveness of the sliding surface:
\begin{equation}\label{eq:control}
    u = - \hat{M}_s^{-1} [\hat{f}_s + \hat{d} + \dot{s}_r + \kappa\,\text{sat}(s/\varphi)]
\end{equation}
where $\hat{M}_s$, $\hat{f}_s$, and $\hat{d}$ are, respectively, estimates of $M_s = \gamma[\alpha_a M_{aa}^{\prime -1} - \alpha_u M_{uu}^{\prime -1} M_{au} M_{aa}^{-1}] $, $f_s = \alpha_a M_{aa}^{\prime -1}f_{a}^{\prime} + \alpha_u M_{uu}^{\prime -1} f_{u}^{\prime}$, and $d = \alpha_a M_{aa}^{\prime -1}d_{a}^{\prime} + \alpha_u M_{uu}^{\prime -1} d_{u}^{\prime}$, $\kappa$ represents the control gain, $\varphi$ stands for the width of the boundary layer, and sat($\cdot$) is the saturation function.

Now, a single-hidden layer network with the sliding variable $s$ as the input neuron is adopted to compute $\hat{d}$:
\begin{equation}\label{eq:ann}
    \hat{d} = \bm{w}^\top \bm{\psi}(s)
\end{equation}
where $\bm{w} = [w_1 \ldots w_n]^\top$ is the weight vector, $\bm{\psi}(s) = [\psi_1 \ldots \psi_n]^\top$ represents the vector with the activation functions $\psi_i$, $i = 1, \ldots, n$, and $n$ is the number of neurons in the hidden layer. It should be noted that if the six state errors ($\tilde{\theta},\dot{\tilde{\theta}},\ddot{\tilde{\theta}},\tilde{\phi},\dot{\tilde{\phi}},\ddot{\tilde{\phi}}$)
have been adopted as input, instead of only $s$, the computational complexity of the neural network would exponentially grow from $n$ to $n^6$.

The chosen ANN can perform universal approximation \cite{Scarselli1998}, hence it can estimate $d$ with an arbitrary degree of accuracy $\varepsilon$, i.e.\ $d = \hat{d}^\ast + \varepsilon$, with $\hat{d}^\ast$ being the estimate related to the optimal weight vector $\bm{w}^\ast$.

Since the adopted sliding surface is a stable manifold \cite{Ashrafiuon2008}, the exponential convergence of the proposed intelligent controller can be proved by means of a Lyapunov-like stability analysis.  Thus, let a positive-definite function $V$ be defined as
\begin{equation}\label{eq:lyap}
    V(t) = \frac{1}{2} s^2 + \frac{1}{2\nu} \bm{\delta}^\top \bm{\delta}
\end{equation}
with $\nu$ being a strictly positive constant and $\bm{\delta} = \bm{w} - \bm{w}^\ast$.

Considering that $\dot{\bm{\delta}} = \dot{\bm{w}}$, the time derivative of $V$ becomes
\begin{equation}\label{eq:lyapdot1}
\begin{split}
    \dot{V}(t) & = s \dot{s} + \nu^{-1} \bm{\delta}^\top \dot{\bm{w}} =  [\alpha_a \dddot{\theta} + \alpha_u \dddot{\phi} + \dot{s}_r]s + \nu^{-1} \bm{\delta}^\top \dot{\bm{w}}\\
               & = [\alpha_a M_{aa}^{\prime-1}(f_{a}^{\prime} + \gamma u + d_a^{\prime}) + \alpha_u M_{uu}^{\prime-1}(f_{u}^{\prime} -\\
               & \quad\gamma M_{au} M_{aa}^{-1} u + d_u^{\prime}) + \dot{s}_r]s + \nu^{-1} \bm{\delta}^\top \dot{\bm{w}}\\
               & = [f_s + d + \dot{s}_r + M_s u]s + \nu^{-1} \bm{\delta}^\top \dot{\bm{w}}\\
               & = [\hat{d}^\ast + \varepsilon - \hat{d} - \kappa\,\text{sat} (s/\varphi)]s + \nu^{-1} \bm{\delta}^\top \dot{\bm{w}}\\
               & = [\varepsilon - \kappa\,\text{sat} (s/\varphi)]s + \nu^{-1} \bm{\delta}^\top [\dot{\bm{w}} - \nu\,s\,\bm{\psi}]
\end{split}
\end{equation}

Since $\text{sat}(s/\varphi)=\text{sgn}(s)$ outside the boundary layer, by updating $\bm{w}$ according to 
\begin{equation}\label{eq:learn}
\dot{\bm{w}} = \nu\,s\,\bm{\psi}
\end{equation}
and defining the control gain as $\kappa > \eta + \varepsilon$, with $\eta$ being a strictly positive constant, we get
\begin{equation}\label{eq:lyapdot3}
    \dot{V}(t) = - [\kappa\,\text{sgn} (s) - \varepsilon]s \leq -\eta|s| 
\end{equation}

Integrating both sides of~\eqref{eq:lyapdot3} yields
\begin{equation}\label{integration}
    \lim_{t \rightarrow \infty} \int_0^t \eta |s| d\zeta \leq \lim_{t \rightarrow \infty} [V(0) - V(t)] \leq V(0) < \infty
\end{equation}

Thus, by evoking Barbalat's Lemma \cite{Hou2010}, it follows that the proposed intelligent controller ensures the boundedness of the tracking error, i.e.\ $(\bm{\tilde{q}},\bm{\dot{\tilde{q}}})\to\Phi$, with $\Phi=\{(\bm{\tilde{q}},\bm{\dot{\tilde{q}}})\in\mathbb{R}^{2m}\vert\:\vert s\vert\le\varphi\}$.

\section{Experimental results}\label{sec:results}
\begin{figure}[h]
\centering{
\includegraphics[width=0.45\textwidth]{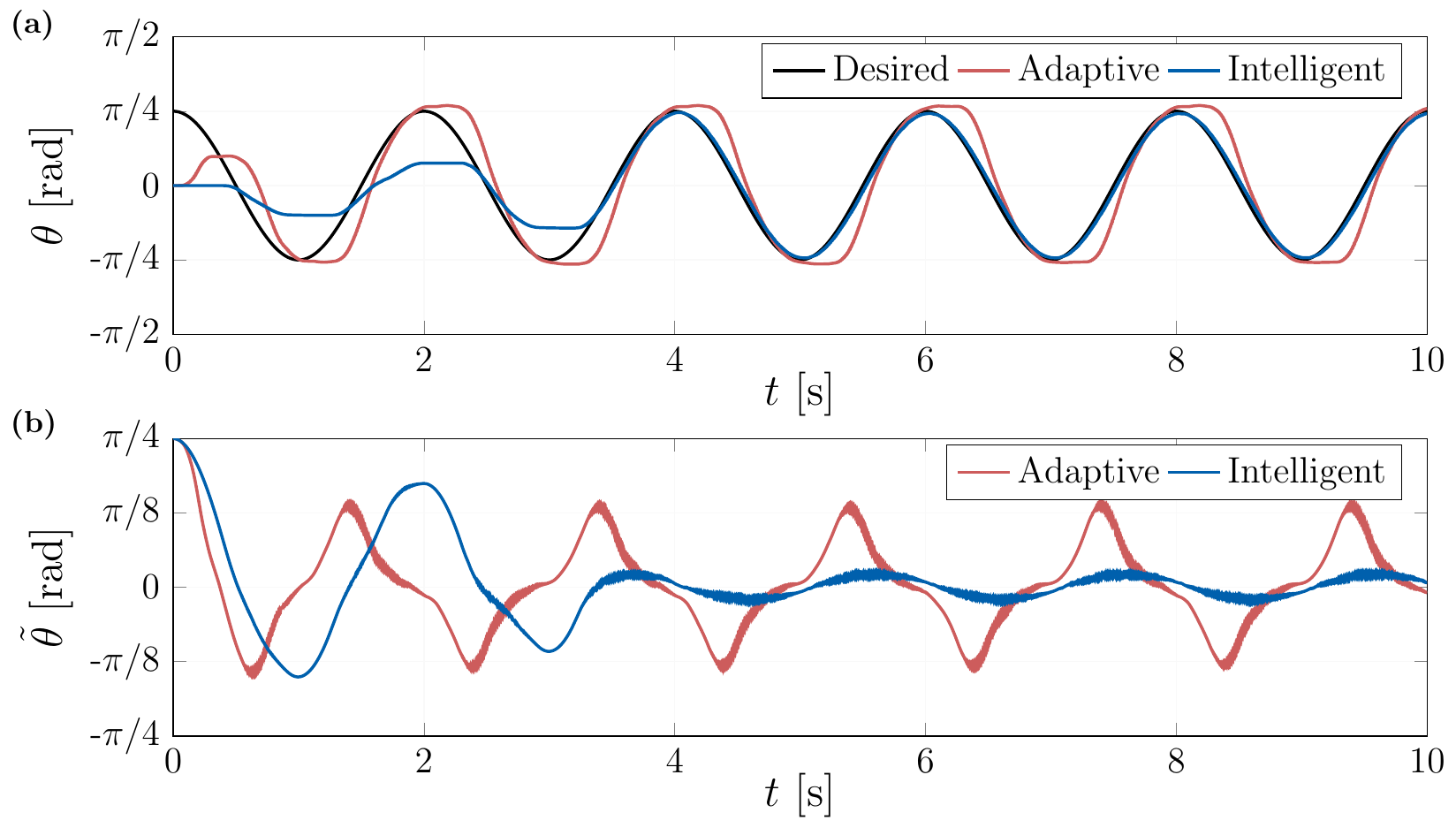}
}
\caption{Experimental results: (a) hub angle and related (b) tracking error.}\label{fig:exper1}
\end{figure}

\begin{figure}[h]
\centering{
\includegraphics[width=0.45\textwidth]{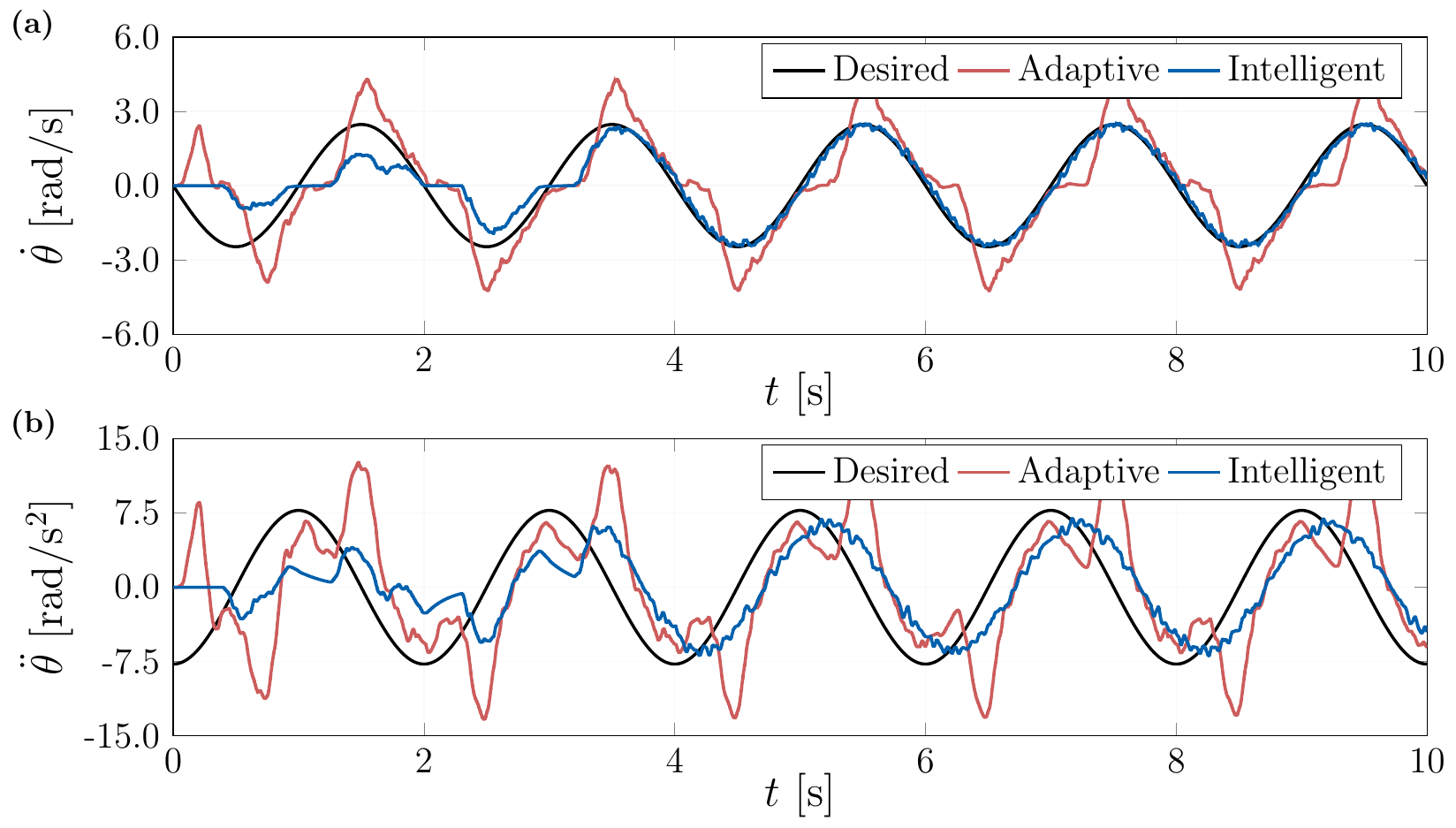}
}
\caption{Experimental results: hub (a) velocities and (b) accelerations.}\label{fig:exper2}
\end{figure}

\begin{figure}[h]
\centering{
\includegraphics[width=0.45\textwidth]{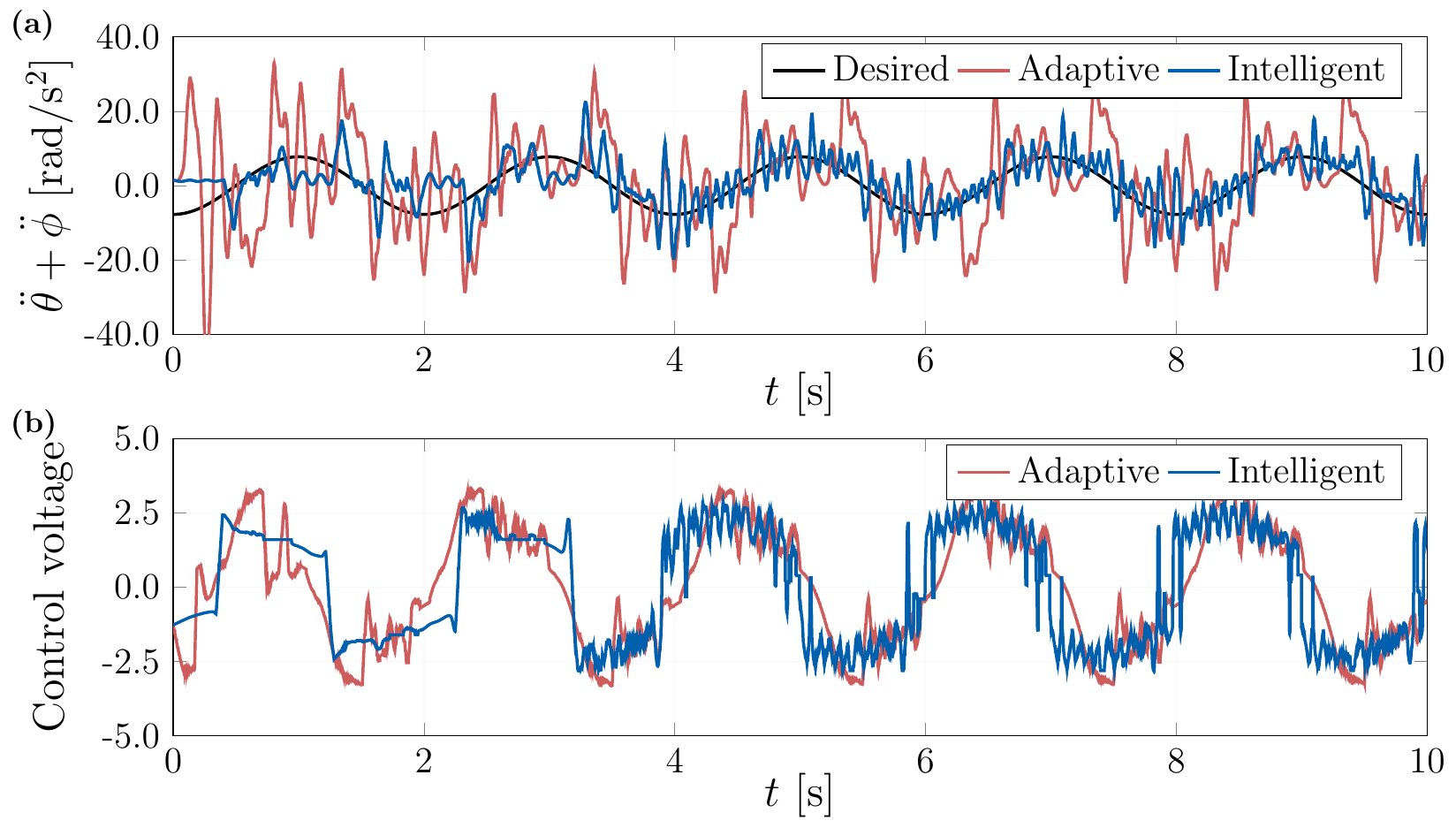}
}
\caption{Experimental results: (a) tip acceleration and (b) control signal.}\label{fig:exper3}
\end{figure}

The effectiveness of the proposed intelligent controller is now evaluated in the experimental setup shown in Figure~\ref{fig:setup}. 
The control scheme was implemented using LabVIEW and myRIO (a real-time embedded device), both provided by National Instruments. The computer code deployed in the experiments, including all the required control parameters, as well as a detailed description of the proposed algorithm and the data used to plot the graphs can be accessed at \url{https://github.com/RoboteamUFRN/Intelligent-Control-Flexible-Manipulator}. 
In order to assess the controller's robustness, as well as its ability to learn online how to compensate for unmodeled dynamics, it is assumed that there is no prior knowledge of the manipulator's stiffness and damping properties, $\hat{f}_s=0$. Moreover, only a rough estimate of the manipulator inertia is considered, $\hat{M}_s = 12.5$. Regarding the adaptive neural network, seven neurons with Gaussian activation functions are adopted in the hidden layer. The weight vector is initialized as $\bm{w}=\bm{0}$ and updated by applying the Euler integration method to~\eqref{eq:learn}.
For a fair evaluation of the adopted method, the proposed intelligent control scheme is also compared with an adaptive sliding mode controller. The intelligent scheme can be easily converted to an adaptive one by setting $\hat{d}_{k+1} \leftarrow \hat{d}_k + \nu s \Delta t$, where $\Delta t$ is the sampling period, $\hat{d}_k$ stands for the estimation at the $k^{\text{th}}$ step and $\hat{d}_0 = 0$. The goal is to allow the manipulator to track a desired trajectory and avoid tip vibrations. Figures~\ref{fig:exper1}--\ref{fig:exper3} show the experimental results obtained with both intelligent and adaptive controllers by tracking $\theta_d = \pi/4\cos(\pi t)$. 

As observed in Figure~\ref{fig:exper1}, the intelligent controller shows a strongly improved performance by allowing the manipulator to track the desired trajectory with a much smaller tracking error. By means of the Integral Time-weighted Absolute Error (ITAE), it can be verified that the control error related to the proposed intelligent controller (ITAE$_{\text{Int}} = 3.44$) is almost 60\% lower than that obtained with the adaptive scheme (ITAE$_{\text{Ada}} = 8.20$).

Considering that both control schemes also require the hub angular velocities and accelerations, sliding mode observers \cite{Shtessel2014} have been used to estimate these variables. Figure~\ref{fig:exper2} shows the obtained estimates, which in fact also evince the better performance of the intelligent controller.

Figure~\ref{fig:exper3} shows the absolute tip accelerations, i.e.\ $\ddot{\theta}+\ddot{\phi}$, and the control signal. Comparing Figures~\ref{fig:exper2}(b) and~\ref{fig:exper3}(a), it can be seen that the high-frequency oscillations come from the measurements with the MMA8452 accelerometer at the tip. 
In fact, this is due to the high noise level associated with low-cost MEMS sensors. Even though this noise could have been filtered out, we chose not to do it in order to avoid unwanted delays and loss of information. It should be highlighted that, regardless of the resulting noise in the control signal, Figure~\ref{fig:exper3}(b), the proposed intelligent scheme is able to deal with it, without degrading its tracking performance, as can be seen in Figure~\ref{fig:exper1}.
Despite the quite reasonable results, replacing the MMA8452 accelerometer with a more accurate sensor is a forthcoming step.

For a better glimpse of the controller's performance, a short video of the experiment can be accessed at \url{https://youtu.be/RIXIVRv5-98}.

\section{Conclusion}\label{sec:conclusion}
This letter introduces a new intelligent controller for single-link flexible manipulators. 
By incorporating an artificial neural network into a sliding mode controller for underactuated mechanical systems, the resulting scheme is able to ensure accurate trajectory tracking, even in the presence of a high level of uncertainty, noisy signals and passive (unactuated) degrees of freedom. In fact, since the weights of the neural network are adjusted online, it can continuously approximate the plant dynamics. Moreover, the adoption of the sliding variable as the input to the neural network exponentially reduces the computational complexity of the control law, allowing its deployment in compact devices such as  low-power microcontrollers. As the control law only requires information regarding the hub angle and tip acceleration, it avoids the need for more sensors along the link. The boundedness and convergence properties of the closed-loop signals are analytically proved by means of a Lyapunov-like stability analysis. The effectiveness of the proposed approach is experimentally validated with a small single-link flexible manipulator, showing a significantly improved performance when compared to a nonlinear adaptive controller.

\begin{acks}
The authors gratefully acknowledge the support of the Brazilian Coordination for the Improvement of High Education Personnel (CAPES) and the German Research Foundation (DFG) by funding the Project ``Manufacturing system models for Industry 4.0 based on highly heterogeneous and unstructured data sets'', PIPC 8881.473092/2019-1 (DFG Grant Number AU 185/72). This work has been also supported by the Brazilian National Council for Scientific and Technological Development (CNPq).
\end{acks}

\balance

\end{document}